\documentclass{article}
\usepackage[preprint]{neurips_2024}
\usepackage{graphicx}

\usepackage[utf8]{inputenc} 
\usepackage[T1]{fontenc}    
\usepackage{hyperref}       
\usepackage{url}            
\usepackage{booktabs}       
\usepackage{amsfonts}       
\usepackage{nicefrac}       
\usepackage{microtype}      
\usepackage{xcolor}         

\title{Weights Augmentation: it has never ever ever ever
let her model down}

\author{$Junbin\ Zhuang^{1}$, $Yan\ Zheng^{2}$, $Baolong\ Guo^{1}$ \& $Yunyi\ Yan^{1,3}$ \\
1. School of Aerospace Science And Technology, Xidian University, China  \\
2. College of Intelligent Systems Science and Engineering, Harbin Engineering University, China \\
3. Corresponding author
}

\begin{document}

\maketitle

\begin{abstract}
Weights play an essential role in deep learning network models. Unlike network structure design, this article proposes the concept of weight augmentation, focusing on weight exploration. The core of Weight Augmentation Strategy (WAS) is to adopt random transformed weight coefficients training and transformed coefficients, named Shadow Weight(SW), for networks that can be used to calculate loss function to affect parameter updates. However, stochastic gradient descent is applied to Plain Weight(PW), which is referred to as the original weight of the network before the random transformation. During training, numerous SW collectively form high-dimensional space, while PW is directly learned from the distribution of SW instead of the data. The weight of the accuracy-oriented mode(AOM) relies on PW, which guarantees the network is highly robust and accurate. The desire-oriented mode(DOM) weight uses SW, which is determined by the network model's unique functions based on WAT's performance desires, such as lower computational complexity, lower sensitivity to particular data, etc. The dual mode be switched at anytime if needed. WAT extends the augmentation technique from data augmentation to weight, and it is easy to understand and implement, but it can improve almost all networks amazingly. Our experimental results show that convolutional neural networks, such as VGG-16, ResNet-18, ResNet-34, GoogleNet, MobilementV2, and Efficientment-Lite, can benefit much at little or no cost. The accuracy of models is on the CIFAR100 and CIFAR10 datasets, which can be evaluated to increase by 7.32\% and 9.28\%, respectively, with the highest values being 13.42\% and 18.93\%, respectively. In addition, DOM can reduce floating point operations (FLOPs) by up to 36.33\%. The code is available at https://github.com/zlearh/Weight-Augmentation-Technology.
\end{abstract}

\section{Introduction}
Deep learning with data augmentation(DA) \cite{he2019bag,dosovitskiy2020image,cubuk2020randaugment,tian2020improving,zoph2020learning} has achieved great success, with preprocessing methods consisting of rotation, translation, scaling, and random cropping \cite{moreno2020improving,shorten2021text,yang2022image}. DA's central core purpose is to make the distribution of the original data set more consistent with the distribution of natural scenes, thereby increasing the diversity and quantity of data. Wang et al. \cite{hao2020improved} improved the Mosaic data augmentation algorithm. After analyzing the synthetic image area, a certain number of training set images are randomly filled in, further enhancing the synthesis ability. CMU \cite{trabucco2023effective} proposed the DA-Fusion strategy using pre-trained text-to-image. Diffusion models generate variants of real images, effectively improving data diversity.

\begin{figure}[!t]
\centering
\includegraphics[width=3.5in]{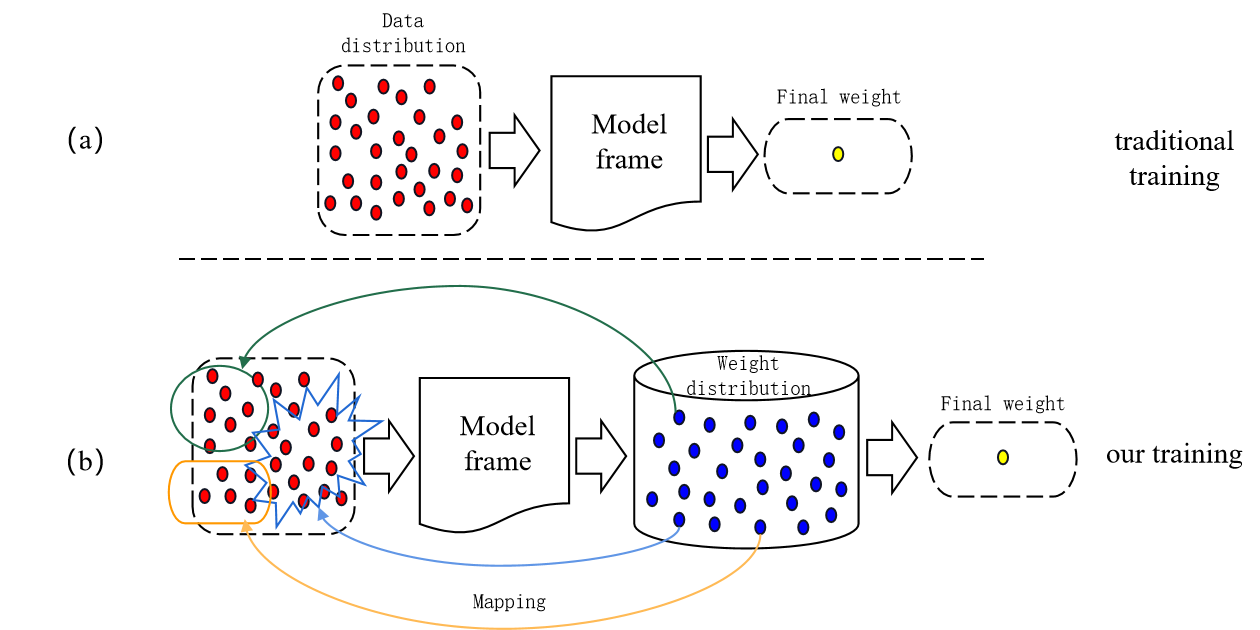}
\caption{We introduce an innovative model training strategy that does not learn the weights directly from the data but gains the distribution of weights through the data. Subsequently, weights for inference are learned from the distribution. 
Red circles denote the data points, blue indicates the weights developed during training, and yellow signifies the final weights obtained. (a) is a traditional training method where data is used to obtain specific parameters. In contrast, (b) illustrates our novel training method. Here, Arrows of different colors indicate that this weight has a better processing effect on the data (corresponding shape). Finally, the final weight is obtained by learning the weight distribution.}
\label{fig_1}
\end{figure}

Weight is the core part of model and the basis for model decision-making in practical applications \cite{zamfirescu2023johnny,alberts2023large,ghosal2023text}. Izmailov et al. \cite{izmailov2018averaging} introduced Stochastic Weight Averaging, which averages weights during training rather than relying on the final set. This is necessary because finding parameters that map well to high-dimensional feature spaces is challenging, and inference requires a single set of weights.
Figure \ref{fig_1}.a is the traditional training method, in which data and weights are many-to-one. Model training aims to find the weights that match the data. The formula can be expressed as:
\begin{equation}
\Phi : \mathbb{R}^{m} \rightarrow x
\end{equation}
Here, $\Phi$ denotes the feature and single-weight relationship mapping. $\mathbb{R}^{m}$ represents the space of feature vectors, the m-dimensional real vector space. Similarly, 
$\mathbb{R}^{n}$ signifies the space of potential weight vectors, the n-dimensional vector space.
\begin{figure}[!t]
\centering
\includegraphics[width=3.5in]{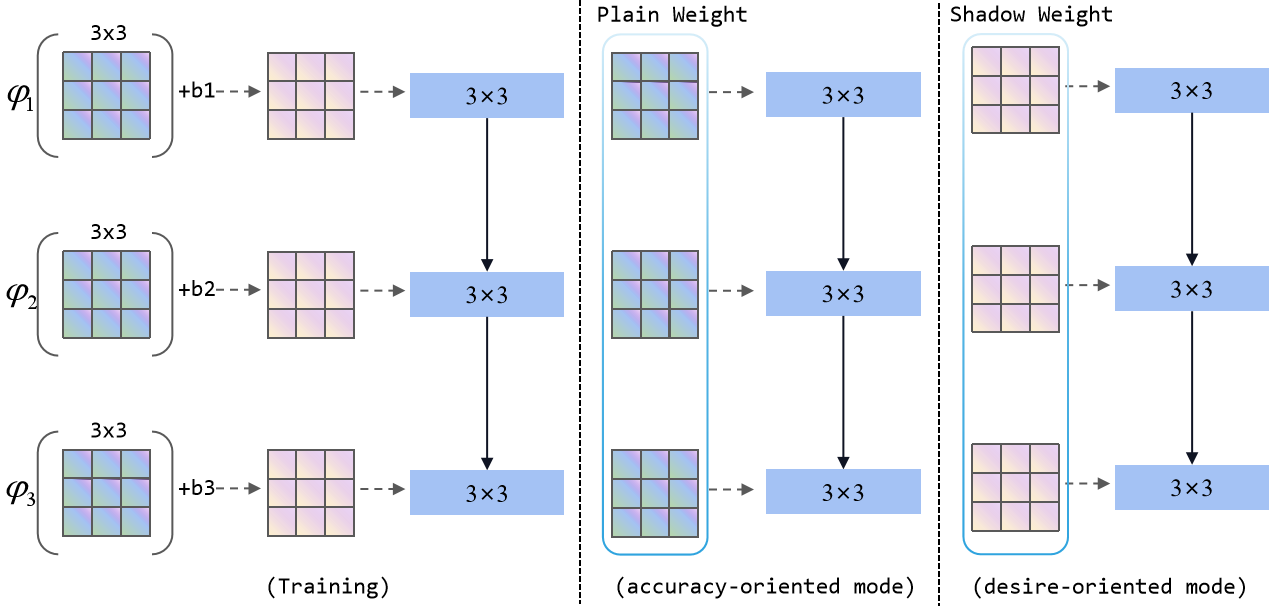}
\caption{Sketch of WAS architecture. There are two modes of inference.  Accuracy-oriented mode
 uses Plain Weight(PW) and desire-oriented mode uses Shadow Weight(SW). Here we only show a part of the network. As inspired by data augmentation, we also use WAS, but only for training.}
\label{fig_2}
\end{figure}
Figure \ref{fig_1}.b is our new strategy. We think the relationship between data and weights is many-to-many. As training, the weight space will become more and more complete. When WAS is used, the weight space will be expanded again. However, we do not find the weights that best meet the requirements of the weight space; we indirectly learn the distribution of the weight space. The announcement is as follows:
\begin{equation}
\Psi:\mathbb{R}^{m} \rightarrow \mathbb{R}^{n}
\end{equation}
where $\Psi$ represents the mapping between feature space and weight space.

Dropout \cite{srivastava2014dropout,bouthillier2015dropout,tobergte2013improving} addresses overfitting by randomly dropping training units, boosting model robustness. It doesn't reduce computational complexity during inference, and there are alternative methods to unit dropping \cite{suresh2021adversarial,zheng2024toward,wang2022makes,jiang2020can,gou2021knowledge}. Influenced by dropout and data augmentation, we propose Weight Augmentation Technology. The core of WAS is to transform weights during training randomly. We call SW that randomly transformed weight. The model uses SW to calculate the loss function to affect parameter updates. However, stochastic gradient descent (SGD) works on Plain Weight(PW). The model can achieve better results only when the actual weight is compatible with most shadow weights simultaneously. 

Although there are two different types of weights, we only need to save PW, and then PW produces SW through random transformation. By this way, we can achieve weights with two modes with different functions at a minimal cost. During task execution, the working mode can be switched according to needs. As shown in Figure \ref{fig_2}, The weight of accuracy-oriented mode uses PW, which is model accuracy-oriented. It can handle most application cases. The weight of the desire-oriented mod uses SW, which can grant the model extra desired properties, such as lower computational complexity, low sensitivity to special data, etc.  

Our contributions are summarized as follows: (1)We propose WAS, which performance of model training by random transformations to change weights.
(2)The weight dual-mode concept is proposed. In inference, only one model weight needs to be retained to achieve multiple states of the model and cope with the needs of different tasks. (3)Negative attitude towards traditional training methods that use data training to obtain a weight. Through train, what is received is the distribution of weights, not a weight. And our ultimate goal is to find the most robust weight in this distribution.
\section{Related work}
With the continuous development of deep learning, researchers have begun to seek more efficient and economical methods to obtain model weights. WiSE-FT \cite{wortsman2022robust} enhances the robustness of fine-tuned pre-trained models by integrating the weights of zero-shot and fine-tuned models. This method can maintain high accuracy and perform well when the distribution changes. However, the performance of WiSE-FT depends on the pre-trained model and fine-tuning data. Guo et al. \cite{guo2020online} demonstrated the Knowledge Distillation method via Collaborative Learning, which trains multiple student models simultaneously during training, allowing the student models to learn collaboratively to enhance their performance. This eliminates the need for an additional teacher model. However, the choice and number of student models can affect the final inference results, and this method has higher hardware requirements. Zhang et al. \cite{zhang2019your} extended Self Distillation, which uses the network as a student and a teacher model and helps transfer knowledge within the network. It does not require an additional pre-trained teacher model and can improve accuracy without increasing inference time. However, Self Distillation introduces an additional shallow classifier, which prevents the model convergence and increases the complexity of training. 

Ensembles \cite{valentini2002ensembles, hansen1990neural,bachman2014learning} can combine multiple weak neural networks to improve model performance. As the number of networks increases, the cost of the ensemble increases linearly during training. Wen et al. \cite{wen2020batchensemble} showed that BatchEnsemble has lower computational and memory costs than typical ensemble methods. It alleviates the fatal shortcomings of ensemble methods, but the memory cost is proportional to the number of neural networks. The Latest Weight Averaging (LAWA) method introduced by Kaddour et al. \cite{kaddour2022stop} improves the convergence speed when training visual or language models on large datasets by averaging the model weights of the most recent k checkpoints, significantly shortening the training period. To some extent, the performance of LAWA is affected by hyperparameters. Therefore, it is necessary to test hyperparameters for specific tasks and datasets. To address the problem of integrating multiple models under limited resources, Singh et al. \cite{singh2020model} used optimal transport for layer-by-layer fusion of various models, which can achieve knowledge transfer without retraining. Wortsman et al. \cite{wortsman2022model} achieve Model Soups, which averages the weights of multiple fine-tuned models to enhance model performance. It requires the simultaneous training of all models to ensure differences between models. 

To address the shortcomings of the above methods, we propose WAS. We randomly transform PW during training to obtain SW. PW is used to be compatible with various SW, enhancing the network's robustness and reducing the model's sensitivity to noise.

\begin{figure}[!t]
\centering
\includegraphics[width=3.2in]{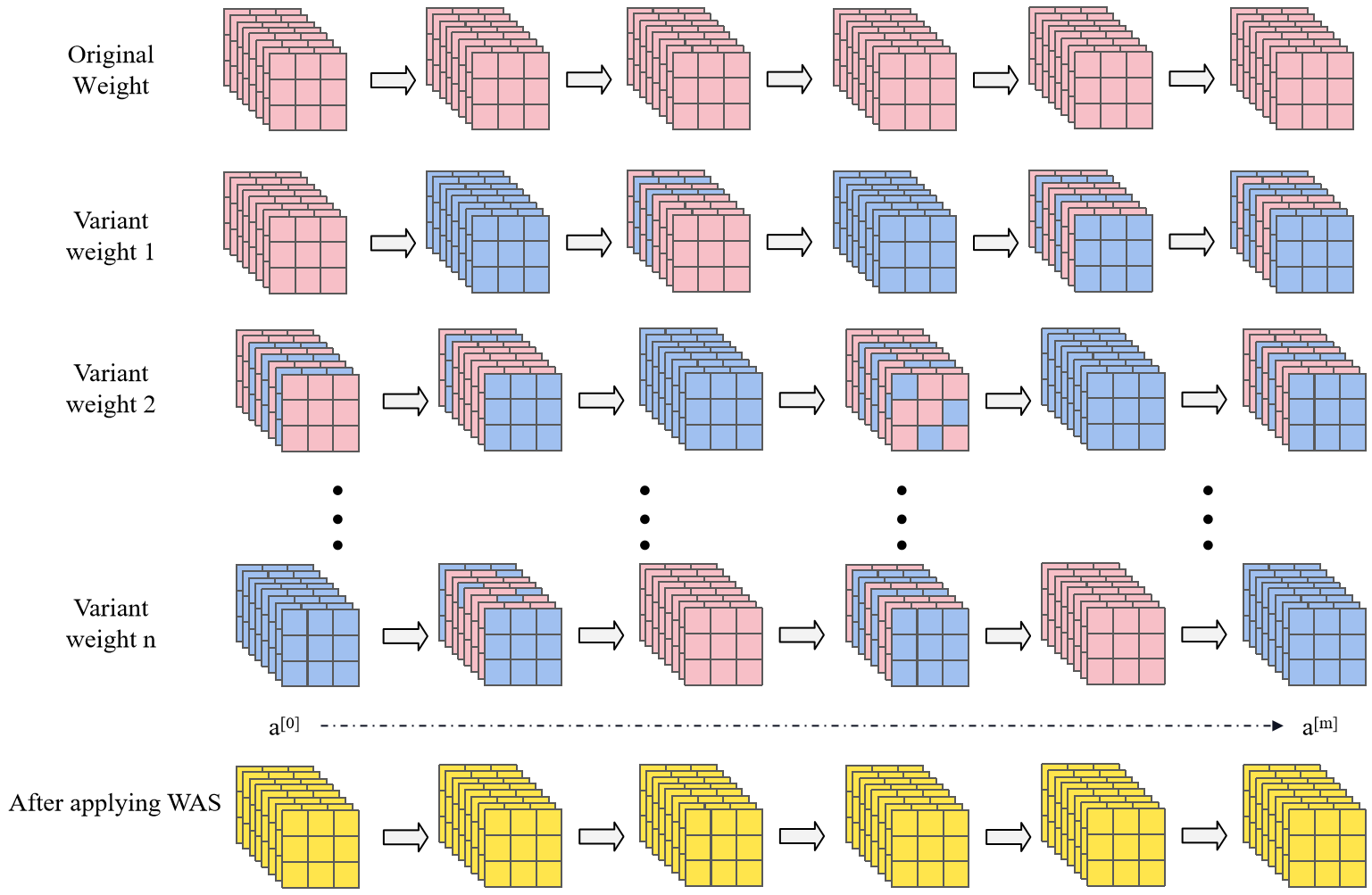}
\caption{WAS is used to generate variant weights.
Here we show the 3x3 convolutional kernel of m layers. WAS enables precise tuning of kernel parameters, allowing adjustments to single or multiple kernels as required. The original weights are represented in red, the variant weights after WAS processing are in blue, and the yellow highlights indicate the weights that are finally determined.}
\label{fig_3}
\end{figure}

\section{Method}
\subsection{Weight augmentation strategy}

Some weights are considered to be "by-products" or "useless", but they can potentially help networks from different perspectives. On specific data, they exhibit good performance but could be more optimal when dealing with broader data. It is discarded during the model selection process. However, with the deepening of deep network research, researchers have begun to recognize the potential value of these "by-products". For example, Ensemble Learning \cite{dietterich2002ensemble,krawczyk2017ensemble,huang2009research} can integrate these seemingly useless weights to form a more powerful and robust model. This method takes advantage of different weights on different data and ultimately improves the model's overall performance by combining high-quality weights. This approach is merely the stitching of weights and has yet to reach the level of weight augmentation. A diverse weight space is crucial, and weights learned from weight distributions perform better. The formula for increasing the diversity of the weight space is:
\begin{equation}
h\left (  x \right ) =ReLU\left ( TW^{'}x  \right ) = ReLU\left ( \sum_{i=1}^{k}  T _{i}W _{i}x_{i} +  b \right )
\label{eq_1}
\end{equation}
where $W$ is the model's weight vector. $W^{'}$ is its transpose. ReLU is a popular activation function. $T_i$ represents an augmentation strategy with transformations like rotation, translation, and scaling.
\begin{figure}[!t]
\centering
\includegraphics[width=1.8in]{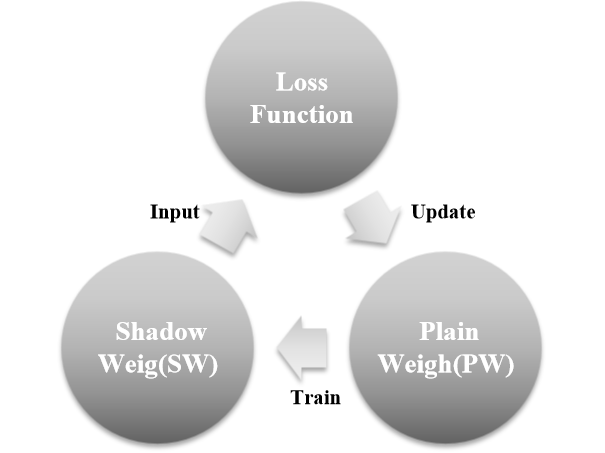}
\caption{WAS  make "process triangle" where the interplay between its elements is evident: SW influences the loss function, which subsequently impacts PW, and PW, in turn, affects SW, creating a cyclical relationship.}
\label{fig_4}
\end{figure}
Reorganizing weights may cause bias and hinder generalization due to incomplete weight distribution capture. To solve this, we introduce randomness, the formula is as follows:
\begin{equation}
h\left (  x \right ) = ReLU\left ( \sum_{i=1}^{j} \gamma  \left ( T _{i} \right ) W _{i}x_{i} +  b \right ) 
\end{equation}
$\gamma$ is significant as it's not defined by a simple function. We enhance randomness through Whether to use WAS or select WAS, etc. 

As shown in Figure. \ref{fig_3}, each execution of mini-batch gradient descent WAS will lead to a weight change, thus producing a variant of SW. As training continues, the distribution of SW gradually improves. The model parameter update formula based on weight update is as follows:
\begin{equation}
\theta_{j}=\theta_{j}-\alpha \cdot \frac{\partial}{\partial \theta_{j}} J(\theta)
\end{equation}
\begin{equation}
\theta_{j}^{pw} =\theta_{j}^{pw} -\alpha \cdot \frac{\partial}{\partial \theta_{j}^{pw}} J(\theta_{j}^{sw})
\end{equation}
Where $\theta_{j}^{sw}$ and $\theta_{j}^{pw}$ represent the jth parameter of SW and the jth parameter of PW, respectively. $\alpha$ denotes the learning rate. $J(\theta)$ is the loss function. $\frac{\partial}{\partial \theta_{j}^{sw}}$ indicate the partial derivative of the loss function $J(\theta_{j}^{sw})$ with respect to the parameter $\theta_{j}^{pw}$. Equation(6) is the formula of the original network Gradient Descent with SGD, and Equation(6) is the SGD formula using WAS. 
\begin{equation}
\frac{\partial J}{\partial \theta _{j}}=\sum_{x,y} \frac{\partial J}{\partial Z_{xy}} \cdots \frac{\partial Z_{x y}}{\partial \theta _{j}}
\end{equation}
where $Z_{x y}$ represent the feature map at position (x, y), $\frac{\partial J}{\partial Z_{xy}}$ is the partial derivative of the loss function relative to $Z_{xy}$, $\frac{\partial Z_{x y}}{\partial \theta _{j}}$ denote the partial derivative of a relative to $\theta _{j}$. We use SW to calculate the loss function to evaluate the difference between output data and the actual results. This evaluation result will directly affect the update of PW. Additionally, SW is used to learn the distribution of data, and we use it as the input of the loss function to approximate the data distribution. PW mainly learns the distribution of SW, so only PW needs to be updated. The above is the biggest difference between this method and the traditional method. The relationship between PW, SW, and loss function is shown in Figure \ref{fig_4}.

In their research, Ding et al. \cite{ding2021repvgg} proposed the Repvgg, which reduces time complexity by adjusting weights during inference while maintaining model performance. This method exploits the inherent potential relationship between the model and the weights but does not associate the model with the weights. In this field, Zheng et al. \cite{zheng2023learn} indicate the "Learn From Model" concept in their paper, emphasizing the importance of research, modification, and design of Foundation models based on model interfaces. Due to researchers' superficial exploration of this field, more research is needed to uncover the intrinsic interaction between weights and model structure. Therefore, although the concept has revolutionary potential, direct manipulation of weights is undoubtedly challenging without such understanding. At this point, if the model is consciously guided during training. Doing so allows human intervention during inference, and direct adjustments to the weights do not cause significant degradation in model performance. To achieve this goal, we will use weight-based training instead of data-based training, allowing the final model to learn from the weight distribution instead of extracting features directly from the data. We can generate a lot of flawed weight through data-driven training, but it can still capture subtle features and deep relationships of data specific to particular datasets. If the model training based on data distribution is regarded as one stage, then the model training based on the distribution of defect weights is called two stages. These two stages alternate with each other in the entire training. The final weights depend on the model's autonomous learning without human intervention. Such a design allows the model to be trained directly to optimize its structure and weights.

WAS encourages the model to explore a broader area of the weight space, promoting high-performance weight configurations while avoiding over-reliance on specific features. It also improves the model's computational efficiency, reducing resource consumption training and inference.

\subsection{Dual working mode of network model}
The core of WAS is to encourage the generation of a large number of weights, which makes the weight distribution more complete, enhancing the generalization and robustness of the model. The model obtained always has two modes in training: AOM and DOM. AOM is akin to conventional deep learning models, as it is directly applied to inference. However, the key innovation of AOM lies in its dynamic weight formation mechanism. This is achieved through a competitive process that unfolds during training. The competition continuously drives the model to optimize its performance. Specifically, SW demonstrates superior performance and will effectively reduce the loss function. As a result, it significantly impacts PW determination. Conversely, those with poorer performance will have less influence on AW determination. 

The second mode focuses on changing WAS, which has solved specific needs such as improving the model's specificity, sparsity, and computational efficiency. In DOM, we can devise WAS tailored to the particular demands of the task and implement it within training. Subsequently, this strategy can be used to tune PW for inference. Although the model's prediction accuracy may be worse than AOM in this mode, it has significant advantages in specific tasks. For instance, random weight cropping can increase the matrix's sparsity of weight. This approach can reduce computational load, potentially by several orders or even tens of times, which is particularly advantageous for applications involving large-scale data processing or model deployment within environments with constrained resources. In summary, the DOM approach can reduce computational, potentially by several or even tens of times, which is particularly advantageous for applications involving large-scale data processing or model deployment within Special needs.

Through WAS, we have implemented a flexible model management strategy. In inference, only one weight needs to be saved to realize two different weight functions simultaneously, and the working mode can be flexibly switched according to the application scenario. This dual-mode simplifies the management of model storage and maintenance. It also strengthens the algorithm's resilience and adaptability.
\begin{table}
  \centering
\caption{Comparison of top-1 accuracy (\%) and average FLOPs (M) on CIFAR10 and CIFAR100. Model-Cand Model-CT denotes models utilizing WAS for random Crop, random Translation, and both random Crop and Translation.}
\label{Table2}
\scalebox{0.5}{
\begin{tabular}{ccccccc}
    \toprule
    \multicolumn{1}{c}{\textbf{Model}} & 
    \multicolumn{2}{c}{\textbf{CIFAR10}} & 
    \multicolumn{2}{c}{\textbf{CIFAR100}} & 
    \multicolumn{1}{c}{\begin{tabular}[c]{@{}c@{}}\textbf{Average}\\ \textbf{FLOPs (M)}\end{tabular}} \\
    \cmidrule{2-3} \cmidrule{4-5}
    & AOM Top1 & DOM Top1 & AOM Top1 & GOM Top1 & \\
    \midrule
    \textbf{VGG} & 87.42 & - & 59.01 & - & 333 \\
    VGG16-C & 89.29 & 88.95 & 62.19 & 60.53 & 321 \\
    VGG16-CT & 91.15 & 90.60 & 63.53 & 61.96 & 279 \\
    \midrule
    \textbf{ResNet18} & 85.53 & - & 59.70 & - & 608 \\
    ResNet18-C & 88.15 & 87.38 & 63.42 & 62.44 & 542 \\
    ResNet18-CT & 89.83 & 88.64 & 63.40 & 61.50 & 315 \\
    \midrule
    \textbf{ResNet34} & 86.54 & - & 58.64 & - & 1214 \\
    ResNet34-C & 89.18 & 88.81 & 62.19 & 61.24 & 1099 \\
    ResNet34-CT & 91.71 & 90.73 & 60.64 & 58.73 & 709 \\
    \midrule
    \textbf{GoogleNet} & 90.55 & - & 72.04 & - & 1457 \\
    GoogleNet-C & 91.99 & 91.43 & 73.05 & 72.02 & 1237 \\
    GoogleNet-CT & 92.68 & 92.04 & 73.08 & 72.01 & 772 \\
    \midrule
    \textbf{MobileNetV2} & 73.2 & - & 47.02 & - & 47 \\
    MobileNetV2-C & 82.58 & 82.43 & 55.43 & 54.65 & 45 \\
    MobileNetV2-CT & 83.02 & 82.88 & 55.92 & 53.95 & 37 \\
    \midrule
    \textbf{EfficientNetLite} & 73.05 & - & 43.13 & - & 8.00 \\
    EfficientNetLite-C & 80.96 & 79.97 & 49.08 & 48.47 & 6.37 \\
    EfficientNetLite-CT & 83.72 & 81.23 & 50.31 & 47.41 & 4.66 \\
    \bottomrule
\end{tabular}
}
\end{table}
\section{Experiments}
We conducted ablation studies and comparative analyses to assess the performance of diverse variant networks in the CIFAR-10 and CIFAR-100 datasets \cite{krizhevsky2009learning}. Our primary aim was to highlight the pivotal role that the integration of WAS plays within our models. Furthermore, we evaluate to delineate the distinctions and performance implications of dual operational modes. The findings from these studies consistently affirm the benefits that WAS confer upon enhancing the model. 
\begin{table}
  \caption{Comparison of AOM and DOM Top-1 Drop Rates under Different WAS}
  \label{Table3}
  \centering
  \scalebox{0.5}{
  \begin{tabular}{ccccc} 
    \toprule
    \multicolumn{1}{c}{\begin{tabular}[c]{@{}c@{}}\textbf{Model}\end{tabular} } & 
    \multicolumn{1}{c}{\begin{tabular}[c]{@{}c@{}}\textbf{Data}\\ \textbf{Augmentation}\end{tabular}} & 
    \multicolumn{1}{c}{\textbf{Parameters}} & 
    \multicolumn{1}{c}{\begin{tabular}[c]{@{}c@{}}\textbf{AOM Top-1}\\ \textbf{Drop Rate(\%)}\end{tabular} } & 
     \multicolumn{1}{c}{\begin{tabular}[c]{@{}c@{}}\textbf{GOM Top-1}\\ \textbf{Drop Rate(\%)}\end{tabular} } \\
    \midrule
    & & $(0^\circ,15^\circ)$ & 5.09 & 4.86 \\
    & & $(0^\circ,45^\circ)$ & 39.52 & 25.58 \\
     \textbf{VGG16-R} & rotate & $(0^\circ,90^\circ)$ & 46.29 & 42.74 \\
    $(0^\circ,90^\circ)$& & $(0^\circ,135^\circ)$ & 54.61 & 53.65 \\
    & & $(0^\circ,180^\circ)$ & 58.40 & 57.76 \\
    \midrule
    &  & (10\%,10\%) & 1.72 & 2.33 \\
    \textbf{VGG16-T}&translate & (20\%,20\%) & 6.62 & 8.47 \\
    (30\%,30\%)& & (30\%,30\%) & 14.42 & 18.21 \\
    & & (40\%,40\%) & 25.76 & 31.76 \\
    \midrule
    &  & (0.8,1.0) & 3.34 & 2.79 \\
    \textbf{VGG16-C}&crop & (0.6,1.0) & 6.49 & 6.71 \\
    (0.8,1.0)& & (0.4,1.0) & 13.04 & 12.12 \\
    & & (0.2,1.0) & 24.19 & 22.42 \\
    \bottomrule
  \end{tabular}
  }
\end{table}
\begin{table}[]
\caption{Comparison of AOM and DOM Top-1 drop rates with varying WAS and data augmentation.}
  \centering
\label{Table4}
\scalebox{0.5}{
\begin{tabular}{ccccc}
    \toprule
    \multicolumn{1}{c}{\begin{tabular}[c]{@{}c@{}}\textbf{Model}\end{tabular} } & 
    \multicolumn{1}{c}{\begin{tabular}[c]{@{}c@{}}\textbf{Data }\\ \textbf{Augmentation}\end{tabular}} & 
    \multicolumn{1}{c}{\textbf{Parameters}} & 
    \multicolumn{1}{c}{\begin{tabular}[c]{@{}c@{}}\textbf{AOM Top-1}\\ \textbf{Drop Rate(\%)}\end{tabular} } & 
     \multicolumn{1}{c}{\begin{tabular}[c]{@{}c@{}}\textbf{DOM Top-1}\\ \textbf{Drop Rate(\%)}\end{tabular} } \\
    \midrule
    
  &                              &(0.8,1.0)                 &2.33 &2.50 \\
   & corp                          & (0.6,1.0)               & 5.93  & 5.15  \\
 &                                  & (0.4,1.0)               & 11.35 & 11.70 \\
VGG16-R &                            & (0.2,1.0)               & 22.31 & 23.08 \\ \cline{2-5} 
($0^\circ$,$90^\circ$) & & (10\%,10\%)             & 2.56  & 3.27  \\
 & translate                           & (20\%,20\%)             & 9.96  & 10.39 \\
 &                           & (30\%,30\%)             & 19.95 & 20.45 \\
 &                          & (40\%,40\%)             & 32.45 & 32.71 \\
     \midrule
         &                                      & ($0^\circ$,$15^\circ$)   &5.09 & 5.27 \\
      &                           & ($0^\circ$,$45^\circ$)  & 28.39 & 27.97 \\
        &    rotate                                   & ($0^\circ$,$90^\circ$)  & 46.80 & 47.28 \\
VGG16-T &                             & ($0^\circ$,$135^\circ$) & 55.18 & 56.28 \\
(30\%,30\%) &                           & ($0^\circ$,$180^\circ$) & 58.96 & 59.67 \\ \cline{2-5} 
         &                              & (0.8,1.0)               & 2.69  & 2.61  \\
         & crop                          & (0.6,1.0)               & 5.16  & 6.11  \\
         &                             & (0.4,1.0)               & 10.12 & 12.18 \\
        &                              & (0.2,1.0)               & 21.53 & 25.10 \\ 
    \midrule
                 &                         &($0^\circ$,$15^\circ$) &4.80 & 4.85 \\
                &                    & ($0^\circ$,$45^\circ$)  & 26.68 & 28.14 \\
                &    rotate                     & ($0^\circ$,$90^\circ$)  & 46.10 & 45.59 \\
VGG16-C &                           & ($0^\circ$,$135^\circ$) & 54.84 & 53.04 \\
(0.8,1.0)     &                           & ($0^\circ$,$180^\circ$) & 58.98 & 56.51 \\ \cline{2-5} 
     &                          & (10\%,10\%)             & 2.56  & 3.24  \\
& translate                    & (20\%,20\%)             & 8.50  & 9.87  \\
&                           & (30\%,30\%)             & 17.70 & 18.21 \\
&                         & (40\%,40\%)             & 28.81 & 33.70 \\
    \bottomrule

\end{tabular}
}
\end{table}
\subsection{WAS for Classification}
To explore the effectiveness of WAS, we have chosen six class deep learning architectures as our experimental models: VGG-16 \cite{simonyan2014very}, ResNet18 \cite{he2016deep}, ResNet34 \cite{he2016deep}, GoogLeNet \cite{szegedy2015going}, EfficientNet-Lite \cite{tan2019efficientnet}, and MobileNetV2 \cite{sandler2018mobilenetv2}. As illustrated in Table \ref{Table1}, we have chanced a series of WAS for comparative experiments against the baseline.

We deliberately eschewed the incorporation of additional techniques, primarily to mitigate the impact of extraneous variables. On a solitary GPU, we established a global batch size of 128. We employed the conventional SGD, initializing the learning rate at 0.01. Furthermore, we fine-tuned the SGD optimizer, assigning a momentum coefficient of 0.9, thereby augmenting the model's stability and hastening convergence throughout the training regimen.

 On the CIFAR-10 dataset, AOG for the models demonstrated discernible enhancements. Specifically, the accuracy of VGG16-C and VGG16-CT saw an increase of 2.13\% and 4.27\% relative to the baseline VGG16, respectively. For ResNet18, ResNet18-C and ResNet18-CT attained an accuracy improvement of 3.06\% and 5.03\%. Similarly, the ResNet34-C and ResNet34-CT models realized respective accuracy improvements of 3.05\% and 5.97\%. The GoogleNet-C and GoogleNet-CT models recorded accuracy enhancements of 1.59\% and 2.35\%. Within the MobileNetV2 series, the accuracy for MobileNetV2-C and MobileNetV2-CT marked a significant rise of 12.76\% and 13.42\% over MobileNetV2. Lastly, the EfficientNetLite-C and EfficientNetLite-CT models secured accuracy improvements of 9.52\% and 13.23\%. 

On the CIFAR-100 dataset, WAT also demonstrated enhanced performance. The accuracy of AOG for Models VGG16-C and VGG16-CT exhibited respective increases of 5.39\% and 7.66\% over the base VGG16. Similarly, the accuracy of AOG for Models ResNet18-C and ResNet18-CT recorded increases of 6.23\% and 6.19\% relative to the original ResNet18. Models ResNet34-C and ResNet34-CT demonstrated accuracy of AOG enhancements of 6.05\% and 3.41\% respectively, in comparison to ResNet34. The GoogleNet-C and GoogleNet-CT models noted slight improvements in the accuracy of AOG, with gains of 1.40\% and 1.44\% over GoogleNet. The MobileNetV2-C and MobileNetV2-CT models marked a significant advancement in the accuracy of AOG, showing increases of 17.89\% and 18.93\% over MobileNetV2. Finally, Models EfficientNetLite-C and EfficientNetLite-CT displayed significant accuracy of AOG improvements, with increases of 13.78\% and 18.93\% over EfficientNetLite.

The performance of Model-C and Model-CT under DOM is lower than under AOM. Their accuracy difference remains at about 1\% to 2\%, but DOM's accuracy is still about 1\% to 2\% higher than the baseline model. This shows that both modes can improve the performance of the model. Particularly, in the context of lightweight models, these adventures are particularly prominent. Although computational speed is not our main consideration, it is worth noting that the floating point operations (FLOPs) of the GOMs of Model-C and Model-CT are both lower than their corresponding baseline models. The FLOPs of Model-C decreased by about 5\% to 20\%, while the reduction in FLOPs for Model-CT is more significant, with the maximum reduction reaching up to 47\%.	

\subsection{Characteristic of WAS}
WAS gives the model unique capabilities that can be customized to specific needs inference. These capabilities can include many aspects, such as reducing computational complexity, decreasing sensitivity to specific data, and so on. This functionality be used by switching the weights to SW during inference. It can customizable adjustments based on specific requirements, enabling the model to accommodate unique environmental conditions.

Table \ref{Table3} presents three representative WAS: rotation (the random rotation angle of weights is set between 0° and 90°), translation (the random translation range of weights in horizontal and vertical directions is 0\% to 30\%), and cropping (the random cropping ratio of weights is between 0.8 and 1.0), following the official PyTorch example \cite{imambi2021pytorch}. To verify the efficacy of WAS for processing particular data, we implemented random data augmentation on the test set to assess WAS strategies. After using WAS, the network model will have two working modes: AOM, a weight that does not apply WAS inference; and DOM, a weight that applies WAS inference. Shown in Table \ref{Table3}, when the strategy for WAT training is a random rotation of 0°~90°, AOM outperforms DOM in Top-1 accuracy. Especially when the rotation of data is limited to 0°~45° and 0°~90°, the accuracy loss of DOM compared to AOM is 13.84\% and 3.55\% lower. For other rotations, the accuracy loss of DOM is about 1\% lower than that of AOM. When randomly cropping is the WAS strategy, as the weight cropping ratio increases, the accuracy loss of DOM compared to AOM decreases by 0.58\%, 0.22\%, 0.92\%, and 1.77\%\. We noticed that with the increase of the random cropping ratio, the disparity in accuracy loss between DOM Two and AOM initially contracts and subsequently expands. This fluctuation is mainly due to the increase in the random crop ratio during training, which causes the model performance to decrease when the parameters exceed a certain threshold.
In summary, WAS helps to promote the ability to process specific data. We found that when the WAS is randomly translated, the accuracy loss of DOM is higher than that of AOM, which is inconsistent with our preliminary conclusion. After combining random crops, we think the reduction of model parameters leads to the fitting ability of decline.
\begin{table}[]
\caption{Compare the impact of VGG16 using different randomly cropped WAS on rotated data}
  \centering
\label{Table5}
\scalebox{0.5}{
\begin{tabular}{cccccc}
    \toprule
    \multicolumn{1}{c}{\begin{tabular}[c]{@{}c@{}}\textbf{WAT}\\ \textbf{Strategy}\end{tabular} } & 
    \multicolumn{1}{c}{\begin{tabular}[c]{@{}c@{}}\textbf{AOM}\\ \textbf{Top-1 acc}\end{tabular} } & 
    \multicolumn{1}{c}{\begin{tabular}[c]{@{}c@{}}\textbf{DOM}\\ \textbf{Top-1 acc}\end{tabular} } & 
    \multicolumn{1}{c}{\textbf{Parameters}} & 
    \multicolumn{1}{c}{\begin{tabular}[c]{@{}c@{}}\textbf{AOM Top-1}\\ \textbf{Drop Rate(\%)}\end{tabular} } & 
     \multicolumn{1}{c}{\begin{tabular}[c]{@{}c@{}}\textbf{DOM Top-1}\\ \textbf{Drop Rate(\%)}\end{tabular} } \\
    \midrule
    
     & & &($0^\circ$,$15^\circ$) &4.80 &4.85 \\
                    &    &      &   ($0^\circ$,$45^\circ$) &26.68 &28.14 \\
    Crop (0.8,1.0)  &89.72    &88.95      &    ($0^\circ$,$90^\circ$) &46.10 & 45.59 \\
                    &    &      &     ($0^\circ$,$135^\circ$) & 54.84 & 53.04 \\
                    &    &      &     ($0^\circ$,$180^\circ$) & 58.98 &56.51 \\ 
                        \midrule

    &    &   &     ($0^\circ$,$15^\circ$) &5.58 &5.04 \\
                   &         &        &     ($0^\circ$,$45^\circ$) &27.85 &26.85 \\
    Crop (0.6,0.8)  &90.34        &89.71        &     ($0^\circ$,$90^\circ$) & 46.54 &45.82 \\
                   &      &        &        ($0^\circ$,$135^\circ$) &56.18 &53.37 \\
                  &       &         &      ($0^\circ$,$180^\circ$) &59.98 &56.16 \\ 
                          \midrule
    &     & &   ($0^\circ$,$15^\circ$) &5.79 &4.51 \\
                 &        &      &      ($0^\circ$,$45^\circ$) &27.44 &24.62 \\
    Crop  (0.4,0.6)            &88.54         &87.99      &      ($0^\circ$,$90^\circ$) &45.67 &43.25 \\
                &         &      &     ($0^\circ$,$135^\circ$) &54.66 &51.57 \\
               &         &      &      ($0^\circ$,$180^\circ$) &58.01 &53.99 \\
                                         \midrule
             & & &($0^\circ$,$15^\circ$) &5.96 &4.43 \\
                &         &      &($0^\circ$,$45^\circ$) &27.57 &25.72 \\
    Crop (0.2,0.4)    &87.09         &86.19      & ($0^\circ$,$90^\circ$) &45.85 &44.4 \\
               &         &      & ($0^\circ$,$135^\circ$) &54.52 &52.77 \\
                  &         &      & ($0^\circ$,$180^\circ$) &58.01 &54.88 \\ 
    \bottomrule
\end{tabular}
}
\end{table}
\begin{table}
\caption{Results of VGG16 using different WAS on CIFAR10}
\label{Table6}
  \centering
  \scalebox{0.5}{
\begin{tabular}{ccccc}
    \toprule
\begin{tabular}[c]{@{}c@{}}Cropping \\ parameters\end{tabular} &
  \begin{tabular}[c]{@{}c@{}}Translation \\ parameters\end{tabular} &
  \begin{tabular}[c]{@{}c@{}}DOM\\ Acc\end{tabular} &
  \begin{tabular}[c]{@{}c@{}}FLOPs\\ (M)\end{tabular} &
  \begin{tabular}[c]{@{}c@{}}Average\\ sparsity rate(\%)\end{tabular} \\ 
                    \midrule
(1.0,1.0) & -           & 87.42 & 333    & -     \\
(0.8,1.0) & -           & 88.95 & 320.68 & 3.70  \\
(0.6,0.8) & -           & 89.71 & 259.67 & 22.02 \\
(0.4,0.6) & -           & 87.99 & 223.08 & 33.01 \\
(0.2,0.4) & -           & 86.19 & 212.02 & 36.33 \\
-         & (30\%,30\%) & 90.37 & 230.63 & 30.74 \\
(0.8,1.0) & (30\%,30\%) & 90.60 & 277.42 & 16.69 \\
    \bottomrule
\end{tabular}
}
\end{table}
As depicted in Table \ref{Table4}, we have implemented AD pipeline that random translations within the test dataset. Moreover, the WAT strategy deviates from conventional AD. As the parameter for random translation from 10\% to 40\%, the AOM's Top-1 drop rate surged from 2.56\% to 32.45\%, while the DOM's Top-1 drop rate saw a similar rise, climbing from 3.27\% to 33.70\%.
Furthermore, if the test set is replaced with randomly cropped data, as the cropping ratio is reduced from 1.0 to 0.2, the Top-1 accuracy drop rate for both modes increases. For AOM, the drop rate is from 2.69\% to 21.53\%, whereas for DOM, it ascends from 2.61\% to 25.10\%. DOM experiences a more rapid performance degradation rate than AOM, especially at higher cropping ratios. 
Based on the preliminary analysis of the above data, we can conclude that AOM has stronger generalization when the WAS strategy is inconsistent with the data augmentation strategy.

Similarly, as the rotation angle of the test set increases from 0° to 180°. the WAS strategy is random translation. The drop rate for AOM increases from 5.09\% to 58.96\%. DOM soared from 5.27\% to a remarkable 59.67\%. However, when WAS uses random cropping, the drop rate for AOM increases from 4.80\% to 58.98\%, and for DOM, it increases from 4.85\% to 56.51\%. This is the same conclusion we reached above. when WAS uses random cropping, the drop rate for AOM increases from 4.80\% to 58.98\%; for DOM, it increases from 4.85\% to 56.51\%. Contrary to the above conclusion, the drop rate of AOM under extreme conditions is higher than that of DOM. Contrary to the above conclusion, the drop rate of AOM under extreme conditions is higher than that of DOM.

In order to explore the above phenomenon, Table \ref{Table5} illustrates the models trained with different random cropping ratios as WAS. DA uses random rotation. On the test set, how the training parameters of WAS change, the drop rate of AOM remains relatively stable. 
In contrast, as the random cropping ratio of WAS increases, the Top-1 accuracy drop rates of DOM are 2.47\%, 3.82\%, and 4.02\% lower than AOM respectively. This trend shows that DOM is more effective in processing rotated data.
As the cropping ratio ranges from 0.2 to 0.4, the Top-1 accuracy drop rates between AOM and DOM narrow. This is attributed to the number of decreases in parameters, which impacts model fitting.
In summary, data rotation can lead to information loss, akin to the effects of cropping. Therefore, DOM (weight-based random cropping) has a lower drop rate than AOM.

WAS is not only reflected in its accuracy but also in the sparsity of the model. As shown in Table \ref{Table6}, we take the Crop as an example, using different cropping ratios. When the Cropping parameters are (0.4,0.6), the proportion of 0 elements in the entire model accounts for 33.01\%, and the model's accuracy is 87.99\%, which is close to the base model. When the Cropping parameters are reduced to (0.2,0.6), the accuracy decreases by 1.41\%, but the proportion of 0 elements increases by 3.32\%, and the FLOPs are reduced by 11.06M. When introducing translation parameters (30\%,30\%), even without cropping, the sparsity rate is increased to 30.74\%, floating-point operations(FLOPs) are reduced to 230.63M, and the model's accuracy is improved by 3.26\% compared to the base model. Combining cropping and translation weight pruning can further improve the model's accuracy and reduce computational costs.

With cropping parameters set at (0.4,0.6), the model's sparsity reaches 33.01\% with zero elements, and its accuracy is 87.99\%, nearly closing the base model. When the Cropping parameters are reduced to (0.2,0.6), the accuracy decreases by 1.41\%, but the proportion of 0 elements increases by 3.32\%, and the FLOPs are reduced by 11.06M. Introducing translation parameters of (30\%,30\%) boosts sparsity to 30.74\%, cuts FLOPs to 230.63M, and improve accuracy by 3.26\% over the base model. 
Combining cropping and translation, WAS improves model accuracy and simultaneously decreases computational costs.

\subsection{Limitations}
WAS  as an effective approach for convolutional networks. It is capable of enhancing accuracy without increasing computational complexity. WAS shines in the customization of model training for specific tasks, offering tailored optimization strategies. Yet, despite its considerable benefits, WAS has not gained the same level of popularity as data augmentation practices.
\section{Conclusion}
We introduce WAS as a training method for models. Its core is to equip the model with a dual mode in inference, enabling weights to cater to tasks with varying demands. This innovative design empowers researchers to train weights that are finely tuned to the specific demands. In the case of AOM, WAT can boost model accuracy by up to 18.93\% without incurring additional costs. For DOM, WAT can reduce FLOPs by up to 36.33\% while keeping the accuracy intact.

\bibliographystyle{unsrt}
\bibliography{neurips_2024}

\end{document}